# Evaluation of Image Segmentation and Filtering With Ann in the Papaya Leaf

Maicon A. Sartin[12] and Alexandre C. R. da Silva[2]

[1]Department of Computing, UNEMAT, Colider, MT, Brazil
[2]Department of Electrical Engineering, UNESP, Ilha Solteira, SP, Brazil

*ABSTRACT*

*Precision agriculture is area with lack of cheap technology. The refinement of the production system brings large advantages to the producer and the use of images makes the monitoring a more cheap methodology. Macronutrients monitoring can to determine the health and vulnerability of the plant in specific stages. In this paper is analyzed the method based on computational intelligence to work with image segmentation in the identification of symptoms of plant nutrient deficiency. Artificial neural networks are evaluated for image segmentation and filtering, several variations of parameters and insertion impulsive noise were evaluated too. Satisfactory results are achieved with artificial neural for segmentation same with high noise levels.*

## 1. INTRODUCTION

Nowadays, a large sum of financial resources is destined for application of agricultural inputs by producers. The costs with agricultural inputs is a main factor in the permanent expenses of the production system. The soil degradation and crop can be caused by excess in the application of agricultural inputs. The technology adoption in the precision agriculture can avoid losses caused by wrong handling agricultural. In this way, the use efficient of fertilizers brings improve for both, producers and environment.

A monitoring system suitable of the crop is necessary to achieve a correct decision-making and determine specific amounts of agricultural inputs. The macronutrient detection can provide recognition of a pattern that belong verify the plant status, independent of crop. Depending of the macronutrient, its excess or lack can be causes damage to the human and plant [1].

Several works assess the correlation of the images in the domain spatial and spectral with greenness index or plant chlorophyll, and its relation with macronutrients deficiencies. The leaves observation by spectrometry with reflectance index identify of satisfactory form the alteration of macronutrients values as the N(Nitrogen), P (Phosphorus), K(Potassium), Ca (Calcium) and Mg (Magnesium) [2]. The macronutrient (N) recognition by images are made in [3-5]. The analysis of the colors models, RGB (Red, Green, Blue) and HSI (Hue, Saturation, Intensity), are evaluated and related with nitrogen amount applied in the soybean crop [5]. In [3][4] verify the histogram analysis with nitrogen amount and grayscale image. The authors discuss features color images of the leaf and methods for the digital image processing (DPI). The images acquisition are realized of semi-autonomous form in computer.

     

The measure of the chlorophyll content by leaves images makes possible the monitoring of the plant nutritional status and divide vegetation and background soil or wastes. The separability criterion is made in [6] using the thresholding method by histograms in crop of corn, soybean and wheat.

Artificial neural networks (ANN) are popular in the precision agriculture for images classification by satellite in applications as: soil degradation, crops recognition, correlation with chlorophyll content and others. There works with ANN application in smallest area or directly by plant leaf. Classification of foliar differences for disease recognition in the rubber tree is made in [7]. In [8] makes use of radial basis neural network with excellent results for plants classification with several texture features as shape, vein, color and Zernik moments.

In the DIP have several techniques used in the object selection in a scene. Image segmentation by histograms and thresholding are employed in the object identification with often by facility of implementation. Others methods can be used in the images segmentation for refinement precision as: Otsu method, region Growing, watersheds and ANN.

In images segmentation and DIP one of the main concerns are in the noises images utilization, with lighting nonuniform and variable. There are methods to support the image segmentation depending of the image interference, as the smoothing by filters, edge and line detection, neighbor, morphological operations, arithmetic, logic and others.

This paper will be analyzed the best parameters for ANN configuration in the image segmentation in order of nutrient deficiency by papaya leaf. The noises insertion in the mage is evaluated and compared with traditional methods with linear filters based to average. The Section 2 presents materials and methods used in the work and developed system. The evaluation of the best ANN parameters are performed in Section 3 with basis in Otsu method. The best ANN configuration will be applied for image segmentation and filtering of the papaya. The system with two ANN is compared to classical filtering by average simple and weighted. A big variety of configurations is made with masks, neighbor and erosion. The final considerations of the work are in Section 4.

## 2. MATERIALS AND METHODS

### 2.1. Experiments and ANN

ANN are known by high capacity of generalization and classification of the data in nonlinear domain. The ANN organization is well defined in layers, neurons, weights and activation function (AF). The ability to adaptation and robustness in the pattern recognition are features that differentiate the ANN of others methods. Also, ANN can be easily associated with others techniques of processing images and signals, fuzzy logic, genetic algorithm and others. ANN will be applied in two functionalities, image segmentation and filtering.

ANN application with backpropagation algorithm is used and can be customizable in six aspects: (i) quantities of layers; (ii) quantities of neurons in hidden layer (QNHL); (iii) quantities and type data of inputs; (iv) rate of learning (RL); (v) constant of moment (CM); (vi) stopping criterion; (vii) activation function.

In the Segmentation by ANN was used the Otsu method [9] for acquisition of the desirable images (segmented) necessary to training. Between the several generated images were selected the more suitable for segmentation analysis. A large number of variations between images and parameters applied in the ANN were related for production optimal results. Changing effected in the ANN parameters are listed as follow:

International Journal of Computer Science & Information Technology (IJCSIT) Vol 6, No 1, February 2014

1. The item (ii) alternated between 2, 4, 5 and 8;
2. The item (iii) have inputs differentiated with and without filters application in the input image, RGB image, grayscale image, inputs with statistical parameters (ISP), ISP and RGB, ISP and grayscale;
3. The item (iv) was alternated between the values 0.01, 0.1, 0.3 and 0.5;
4. The item (v) was alternated between the values 0.3, 0.5, 0.7 and 0.9;
5. The item (vi) in the first implementations was not alternated, just with 100 epochs, after selection of the best simulations, the process was repeated with 500 and 5.000 epochs.

The development and implementation of three stages have goal to define the optimal parameters for system organizations. A large evaluation was performed in the changing in this parameters. By acquired results can determine a suitable structure to ANN. The system organization and execution of general form in the ANN is illustrated in Figure 1 with selection just of segmentation. The fourth stage makes the test for evaluation of the ANN robustness.

Experiments were performed in high-level language in Matlab without toolbox and functions, except the conversion RGB to grayscale and noise salt and pepper. The images were acquired by a cellphone with 3 Megapixel cam and the resolution was reduced for 320x240 size. Some images with different background and illumination were investigated, but are not specified in this work, because the simulation quantities and training time are elevated. The chosen image of test have a possible nitrogen nutrient deficiency in the plant by symptoms with yellowish colouring found in the leaf [10].

By source image is applied the DIP methods for generate three images types: full, input and desirable. The sequence definition of the methods for each image is described in each stage in Section 3. Full image is used after training or run other method, e.g., a average filter. Input image have the same method of normalization applied in full image, but its size is reduced for approximately 1.82% of the full image, the remaining is used in the validation. Desirable image have the goal of serve as reference for ANN training. This image is generate by insertion of several methods, the main is the Otsu method with multiple thresholds. The method function is segmentation source image and let just image part correspondent to nutrient deficiency, i.e., desirable image.

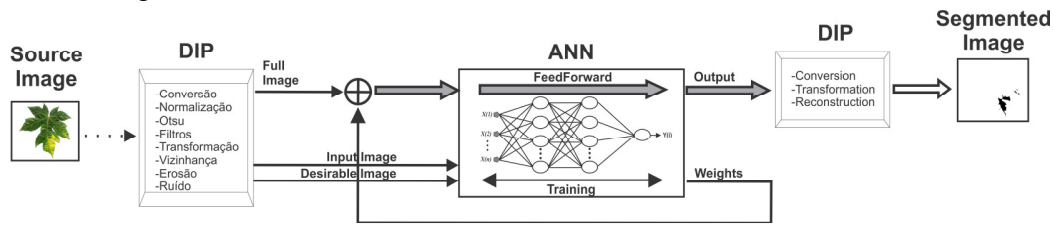

Figure 1. General organization of the ANN system for image segmentation.

Figure 1 correspond the execution of the system of general form. In Figure 2 presents all methods used in the DIP and distinct images defined in each stage. The run sequence of the methods will be defined in each stage in the Section 3.

In training process (13) of the backpropagation algorithm is terminated by stopping criterion, in this case of 100 epochs, with inputs images reduced and desirable. Input image ($f(x,y)$) is included in the ANN after the normalization ($f''(x,y)$) of values in the interval between [0:1]. The finish of the training will generate weights necessary for run the feedforward mode (12) and segment the full image, sequence 1-2, according with nutrient deficiency. The quantity of pixels in the full image is 76.800 and for training is 1.400.

International Journal of Computer Science & Information Technology (IJCSIT) Vol 6, No 1, February 2014

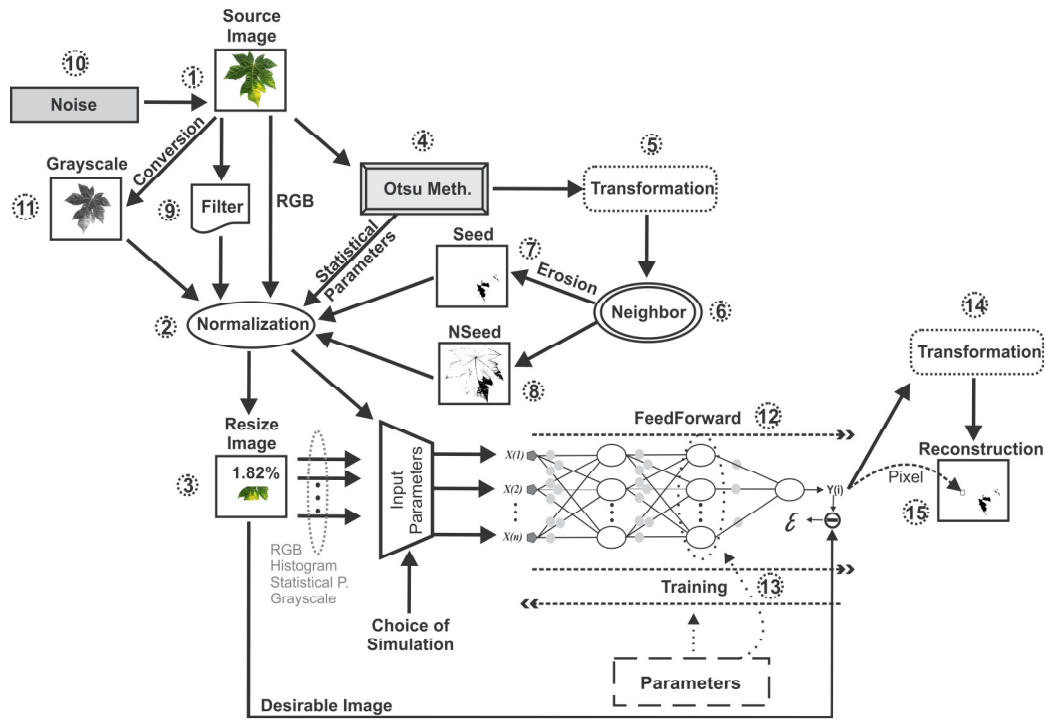

Figure 2. Following implementation of the methods used in all stages for change images.

The output in the Feedforward mode is the pixel of the segmented image, i. e., the running in all simulations in the Section 3 are performed of sequential form pixel by pixel. Output image in all cases yet pass by two process the transformation in grayscale (14) and reconstruction image (15) as a matrix *NxM*. After the two process is obtained a segmented image according with parameters specified in the training and in the ANN structure of each simulation.

## 3. RESULTS AND DISCUSSION

The four stages performed in this work are related, respectively, in Sections 3.1 and 3.2. First stage evaluates the items 1, 3, 4 and 5 in Section 2. The second stage determine the best combinations of linear filters. Third stage changes types and quantities of ANN inputs, item 2. Fourth stage focuses influence of the filters in the segmentation.

### 3.1. Parameters Evaluation

**First stage** consists in the identification of the best parameters between desirable image, RL, CM, and QNHL. In this first stage were performed ANN training and running for two tasks:

1. Variations of Parameters: RL and CM;
2. Change of the QNHL.

The first task have a ANN with 3 neurons in input layer correspondent to components RGB of the image, hidden layer have 4 neurons and output layer with just 1 neuron. Output layer brings the results in a grayscale image to differentiate between two class, with and without nutrient deficiency in the leaf. The AF hyperbolic tangent was used in all neurons and stages.

The images input and desirable are obtained by following sequence, 1-2-3 and 1-4-5-6-7-2-3, respectively, as in Figure 2. The second sequence correspond the image "Seed", as to image "Not Seed" is obtained by sequence 1-4-5-6-8-2-3. In the first task were performed 32 simulations, being 16 with a desirable image based in the Otsu method and erosion operator (Seed) and others 16 with a desirable image based just in the Otsu method (NSeed). The quantity de pixels of the segmented area in the "Seed" image is 2.621 and in "NSeed" is 11.189 pixels.

First 32 simulations were changed to the RL and CM as items (iv) and (v) in Section 2. The first simulations set have goal of analyze the best parameters and desirable image form for ANN. Table 1 presents results of the 6 best simulations.

Learning curves by MSE of the 6 best and five worst simulations are presented in Figure 3. Learning curves of the best simulations in Figure 3-(a) show suitable behavior of the MSE. The simulations 1, 5, 9 and 13 have behavior similar, because the RL is equal and the difference is in the CM. The simulations 2 and 6 have difference more expressive of others, because the RL is larger than the others.

In Figure 3-(b), the learning curves with worst simulations present high values of MSE and some cases exist the growing of the error. The last simulation (16) show a problem of error oscillation caused by high value in the CM [11]. This behaviors unsuitable influence in the results of the ANN, so were not used.

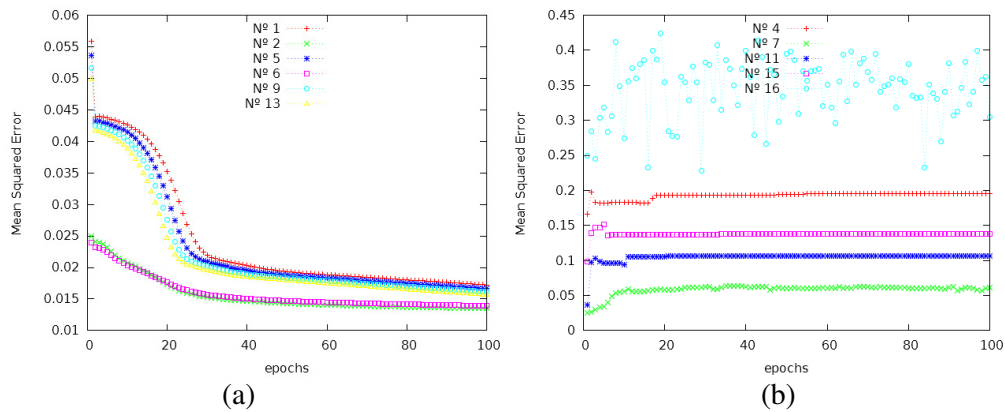

Figure 3. Learning curves of the best and worst simulations with 100 epochs: (a) 6 best; (b) 5 worst.

The best simulations were repeated with 500 and 5.000 epochs. The great difference between training is less than 1% in relation to training with 100 epochs. The results presents in Table 1 are with 100 epochs.

For evaluate the segmentation was made the criterion of percentage of hits and errors (PHE) in each simulation defined by (1) and (2). The calculation is based in the increment of hits and errors according with the suitable limit of gray levels for result.

In the output layer in ANN have a neuron for determine the pattern output or image segmentation. Therefore, the generated image in the output of the ANN and Otsu method are in gray scale. The variable "Hit (H)", in (1), is incremented just when the output belong to close interval between 1/4 of the possibilities of gray levels (L) for above and below of the desirable value. Otherwise, the variable "Error (E)" is incremented. Thus, each pixel is analyzed individually and defining a

International Journal of Computer Science & Information Technology (IJCSIT) Vol 6, No 1, February 2014_

acceptable margin of error. In (1), *D* is desirable value of the pixel for comparison in the outputs and *LMax* is the maximum value for gray level, with 8 bits this value is 255.

$$H \mid E = \begin{cases} H++ & \text{if} \quad D + [(1/4) \times LMax] > \text{Output} > D - [(1/4) \times LMax] \\ E++ & \text{if} \quad \text{Otherwise} \end{cases} \quad (1)$$

$$PH = 100 \times \left[ \sum H / \left( \sum H + \sum E \right) \right] \quad (2)$$

The 6 best simulations have percentage of hits above of 97% in Table 1. The numbers of the simulations (1° column) correspond its correct sequence in the set of 32 simulations effected, all belong to group of images of the type seed.

The mean squared error (MSE) is other parameter for analysis of the validation results with different parts of the image in relation to training. In (3) define the MSE calculation for validation of the ANN. *Y* is ANN output and *D* the desirable value correspondent, and *Pn* is the pixels amount of the image used in the validation of the tests.

$$MSE = \frac{1}{Pn} \left( \sum (Y_i - D_i) \right) \quad (3)$$

Table 1. The 6 best results achieved with changing of the parameters rate of learning (RL) and Constant Moment (CM).

| Simulation | Desirable Image | QNHL | RL | CM | PHE (%) | MSE | Runtime (S) |
|---|---|---|---|---|---|---|---|
| 1 | Seed | 4 | 0.01 | 0.3 | 97.19 | 0.01725 | 208.25 |
| 2 | Seed | 4 | 0.1 | 0.3 | 97.21 | 0.01350 | 207.61 |
| 5 | Seed | 4 | 0.01 | 0.5 | 97.23 | 0.01665 | 204.11 |
| 6 | Seed | 4 | 0.1 | 0.5 | 97.09 | 0.01392 | 204.30 |
| 9 | Seed | 4 | 0.01 | 0.7 | 97.26 | 0.01614 | 204.35 |
| 13 | Seed | 4 | 0.01 | 0.9 | 97.29 | 0.01569 | 218.65 |

In second task was made the neurons variation in the hidden layer with 6 best simulations of the Table 1. The neurons were changed in 5 and 8 generating more 12 simulations, this values were tested because most researchers recommended quantities of neurons (n) in first hidden layer *2xInputs > n > number of inputs*. Thus, were defined one quantity inside of the limit and other out.

The best result was found by simple average between the simulations with different quantities of epochs (100, 500 and 5000) and neurons (4, 5 and 8). The best average was simulation number 13 with RL of 0.01, CM of 0.9 and 4 neurons, as in Table 1. The average percentage obtained was 96.07% and its training parameters will be used in following stages.

**Second stage** have goal of choice the best filters and masks. The two types of linear filters used were by average simple and weighted. In (4) consist in general form of the data computing of the filters *g(x,y)* about a image *f(x,y)* of size *NxM* and mask *W(u,v)*. The constants *p* and *q* delimiting data computing and are found by *p = (k/2)+1/2* and *q = (k/2)-1/2*. For facilitate data indexing were adopted masks with equal orders and odd. In all filters have product between the pixels values of each image component with the value correspond to mask. Simple average filters have

International Journal of Computer Science & Information Technology (IJCSIT) Vol 6, No 1, February 2014

all values of the mask in 1. Weighted average filter the values are decreasing of the center to edge in multiples of 2, the diagonals with values less than orthogonal neighbors to center, i. e., in diamond form [9].

$$g(x,y) = \frac{\sum_{r=p}^{N-q}\sum_{c=p}^{M-q} W(r,c) \times f(x+r, y+c)}{\sum_{r=p}^{N-q}\sum_{c=p}^{M-q} W(r,c)} \quad (4)$$

In the two filters types were alternated between various masks sizes, as in Table 2. For obtain the best filters combinations were used in the simulations the best of the previous stage, number 13 of the Table 1. In this preparation of the input or desirable image for the ANN were generated 144 images according with combination of all possibilities presented in Table 2. By images generated were chosen 13 best combinations for apply in ANN. The 13 simulations achieved results above of 96%, but any result was better than the number 13 in Table 1.

Table 2. Filters Combination applied in the preparation of the desirable image for segmentation and filtering.

| Filter | Mask | Neighbor | Erosion |
|---|---|---|---|
| Simple Average | 3x3, 5x5, 7x7, 9x9, 11x11, 15x15, 35x35 | WN, V4, V8 | WE, 3x3, 5x5, 11x11 |
| Weighted Average | 3x3, 5x5, 7x7, 9x9, 11x11 | WN, V4, V8 | WE, 3x3, 5x5, 11x11 |

**Third stage** were changed the ANN inputs in four combinations: *ANN with 4 inputs* and 5 neurons in hidden layer; and insertion of the histogram as additional value in input; *ANN with 3 inputs:* two statistical values based in the calculation of the Otsu method: accumulated intensity and between-class variance; and third input is a grayscale image; *ANN with 2 inputs:* histogram and a grayscale image; *ANN with 1 input*: a grayscale image.

In this stage were performed 64 simulations, 16 for each combination, with the same changing in stage 1 with just the parameters RL and CM. The four inputs combinations for ANN achieved lower results to presented in Table 1. However, the results obtained show the capacity of the ANN to work with others inputs types and quantities. In all combinations had results above of 90%, at least one configuration. The best results are in first and last combination, respectively, 4 and 1 inputs achieving results above of 97% and 94%. With quantities of inputs smaller had need of training with larger quantities of epochs for stabilize results.

**3.2. Noises analysis**

**Fourth stage** have goal of evaluate the behavior of the ANN and filters with noise insertion in distinct percentages and image components.

The impulsive noise insertion (Salt and Pepper) is a form of validate the robustness of a determined method for this anomaly type. Noises have possibility of occur mainly when is working with images in external environment. Impulsive noise consist in the insertion of intensity values maximum and minimum of random form, respectively, black and white (255 and 0 with intensity levels of 8 bits) in determined component of the image.

This stage are compared the best filters applications in the images for smooth impulsive noises, and were defined three tasks:

1. Insertion impulsive noise with application of the best filters and combinations of techniques for image, ANN with function of image segmentation;
2. Insertion impulsive noise, ANN with function of image segmentation and filtering;
3. Insertion impulsive noise, ANN with function of image segmentation;

The percentages of impulsive noise (Salt and Pepper) in 5%, 10%, 15%, 20%, 25% and 30%. The application of the noise was alternated between each one of the three image components (RGB) and with the three together. Hence, for each simulation will be generated 24 distinct impulsive noise types in the image and without noise. ANN training will be effected with 1.82% of each one of the 25 generated images, 24 with noise and 1without noise.

The ANN parameters will be the same for all simulations and correspond the choice of the Section 3and number 13 in Table 1. In each simulation will be analyzed the hits percentages related to image segmentation and filtering. In image filtering in output will be compared with source image in the three components (RGB), as in (2) and (4).

The limit acceptable for variation of the gray level in the filter is less than 5%. When that in the segmentation the limit is of 50 % as in (1) of the Section 3.1.

$$H \mid E = \begin{cases} H++ & \text{if} \quad D + [(1/40) \times LMax] > Output > D - [(1/40) \times LMax] \\ E++ & \text{if} \quad \text{Otherwise} \end{cases} \quad (5)$$

First task generate 25 images and apply the filter correspondent of the simulation in all images. ANN have function just of segmentation and training is made with 25 images after filter application. Desirable images and five best combinations with filters, masks, neighbor and erosion were found in the second stage. This combinations are used for results production of the first task and comparison with filter application by ANN in the second task.

The second task have just a difference in relation the first task, the first apply several linear filter types in the image with noise for obtaining improve in the images before of the segmentation with ANN. In second task makes use of the owner ANN for filter, the images generated with noise serve of input and source image how goal for ANN training. After ANN training is performed the feedforward mode for produce each filtered image. Training process and run feedforward mode is repeated again, but now the filtered images acquired in the previous process are inputs and segmented image is goal.

This methodology is presented in the results as "ANN-Filter-Segmentation" and the Figure 4 illustrates the process. For ANN determine specific results for each stage were elaborated two ANN, one for filtering and other for segmentation. in filtering ANN is organized in a structure 3-4-3 and segmentation 3-4-1.

The "ANN-WithoutFilter-Segmentation" correspond third task and have not process of image filtering just training with noises images for production segmented image as goal, i. e., just the second process of the second task (Segmentation).

In all results tables have in the simulations of 1 to 5 the full name correspond the combination linear filter used. The part before second hyphen is identification of the filter type and mask size. After following the neighbor type and erosion. The sixth simulation is implementation of the two

International Journal of Computer Science & Information Technology (IJCSIT) Vol 6, No 1, February 2014

ANN, filtering and segmentation. Last simulation have ANN just for segmentation without filter utilization. The relation of the abbreviations with due to simulations are listed follow:

1. Mean-3x3-Neighbor8-Erosion5 (M3N8E5);
2. Mean-5x5-WithoutNeighbor-Erosion11 (M5WNE11);
3. Mean-7x7-Neighbor4-Erosion11 (M7N4E11);
4. Mean-9x9-Neighbor8-Erosion11 (M9N8E11);
5. Mean_Weighted-3x3-Neighbor8-Erosion5 (MW3N8E5);
6. ANN-Filter-Segmentation (AFS);
7. ANN-WithoutFilter-Segmentation (AWFS).

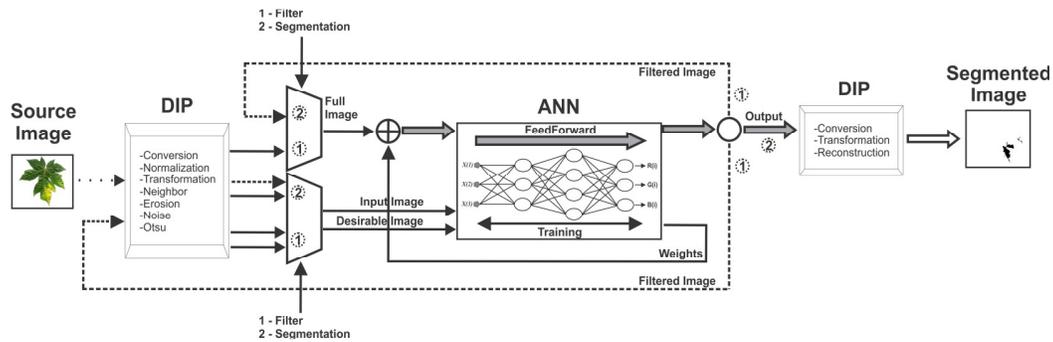

Figure 4. Organization of the implementation of two ANN for filtering and segmentation.

Stage results are contained in 10 tables, 5 for filtering and 5 for segmentation. The first set of simulation is not present the ANN-WithoutFilter-Segmentation (AWFS) by have not contain the filtering process. In the other set of simulation for ANN application with function of filtering and segmentation with results in grayscale. In each set of five tables have results of the noise insertion in each component RGB, and in all components together, and one table with average between the first four.

All results are presented according with percentage of hits correspondent the (1) and (4), respectively, for segmentation and filtering. Tables 4 and 5 present average, respectively, stages of filtering and segmentation. The first column specific the simulation type adopted. The seven following columns are percentage of hits in the noises specified, first without noise and others six with noises of 5% to 30% in the 3 image components. Last column have general average of all simulations, components and noises percentages.

Table 4. Results average of the images filtering in (%).

| Simulation | 0% | 5% | 10% | 15% | 20% | 25% | 30% | Average |
|---|---|---|---|---|---|---|---|---|
| M3N8E5 | 89.23 | 63.87 | 46.66 | 34.67 | 36.06 | 32.14 | 16.85 | 38.37 |
| M5WNE11 | 82.37 | 56.44 | 40.94 | 30.88 | 36.76 | 32.75 | 17.22 | 35.83 |
| M7N4E11 | 77.08 | 57.43 | 40.55 | 29.34 | 36.09 | 32.47 | 15.51 | 35.23 |
| M9N8E11 | 72.62 | 57.25 | 39.25 | 27.29 | 35.22 | 31.10 | 14.26 | 34.06 |
| MW3N8E5 | 91.40 | 67.49 | 50.10 | 38.39 | 38.25 | 33.92 | 19.16 | 41.22 |
| AFS | 71.39 | 69.83 | 68.14 | 66.43 | 68.74 | 67.27 | 62.16 | 67.09 |

The results of the filtering process show that the ANN is better in all cases with noise insertion. In the simulations without impulsive noise is less than others, but maintain the average of the others. With impulsive noise insertion makes general average of the ANN better than any other filter type. The filter of weighted average with mask 3x3 was the simulation with better results compared to other linear filters.

Masks with sizes greater had worst results in its most. The difference between the noises percentage can be observed with expressive form in any cases, same with percentage of 10%. Noise with 30% reduces drastically the percentage of hits in the most of the cases or components, except in simulations with ANN.

In Table 5 presents average found in the segmentation with all percentage of hits, including for image without noise. The results are better than filtering process, due to (1) have more clearance in the precision.

In Table 4 and 5 are presented the results in the employ of ANN for filtering and segmentation with suitable robustness and generalization. In all cases, except number 4 (M9N8E11) obtained results above of 90% same with insertion of 30% noise. In the best simulations the larger difference between image segmentation without noise and with 30% noise was less than 2.5%.

Simulation of number 4 (M9N8E11) has worst results due the greater mask (9x9) between all others linear filters. In the column without noise can observe that is worst value found.

Table 5. Results average of the images segmentation in (%).

| Simulation | 0% | 5% | 10% | 15% | 20% | 25% | 30% | Average |
|---|---|---|---|---|---|---|---|---|
| M3N8E5 | 96.96 | 95.39 | 94.61 | 94.12 | 93.84 | 93.67 | 93.60 | 94.60 |
| M5WNE11 | 98.51 | 98.33 | 98.26 | 97.97 | 97.80 | 97.60 | 97.16 | 97.95 |
| M7N4E11 | 97.85 | 97.76 | 97.72 | 97.68 | 97.61 | 97.46 | 97.14 | 97.60 |
| M9N8E11 | 69.80 | 71.49 | 74.42 | 77.45 | 79.92 | 81.62 | 82.87 | 76.80 |
| MW3N8E5 | 95.81 | 95.53 | 95.16 | 94.89 | 94.61 | 94.43 | 94.10 | 94.93 |
| AFS | 96.86 | 96.30 | 95.77 | 95.28 | 94.77 | 94.28 | 93.82 | 95.30 |
| AWFS | 98.31 | 98.17 | 98.07 | 97.98 | 97.84 | 97.76 | 97.66 | 97.97 |

In the two best results, number 2 (M5WNE11) and 7 (AWFS), have small difference. Number 2 have best results with low insertion noise and number 7 is the opposite. Number 7 have advantage of the a system with just the ANN and number 2 need of a extra module for image filtering before run ANN. Thus, the best case chosen is simulation number 7 (AWFS).

Figure 5 presents result of the image segmentation with larger noises levels for ANN, i. e., the images with noise insertion in all three image components (RGB). The results show capacity suitable of the ANN for goal propose. Figure 5-(c) the segmentation is performed with training of the first stage, number 13 in Table 1, without noise insertion. Figure 5- (d;f;h;j;l;n) are generated images by segmentation with ANN and training with noise insertion. In the comparison of the goal image, Figure 5-(b), the generated image by ANN with and without noise, Figures 5-(c;d), the segmentation by ANN obtained visual results better than the Otsu method.

## 4. CONCLUSIONS

In this paper was presented the evaluation of the best configurations for Artificial neural networks with goal of images filtering and segmentation related the nutrient deficiency in the papaya leaf. Several parameters were analyzed as: RL, CM, QNHL, inputs types and quantities, various filter types and noises insertion.

All parameters analyzed determine the best choice for ANN configuration have following organization: 3 inputs RGB, 3 neurons in the hidden layer, RL of 0.01 and CM of 0.9. The comparison of the ANN with other system with two ANN, one with filter function and other with segmentation, demonstrated that results achieved with just one ANN (optimized) is better than with two ANN for treatment of impulsive noise (Salt and Pepper). ANN optimized obtained better results than with linear filters by average simple and weighted. ANN with few neurons can obtain results satisfactory for image segmentation in the recognition of nutrients deficiency.

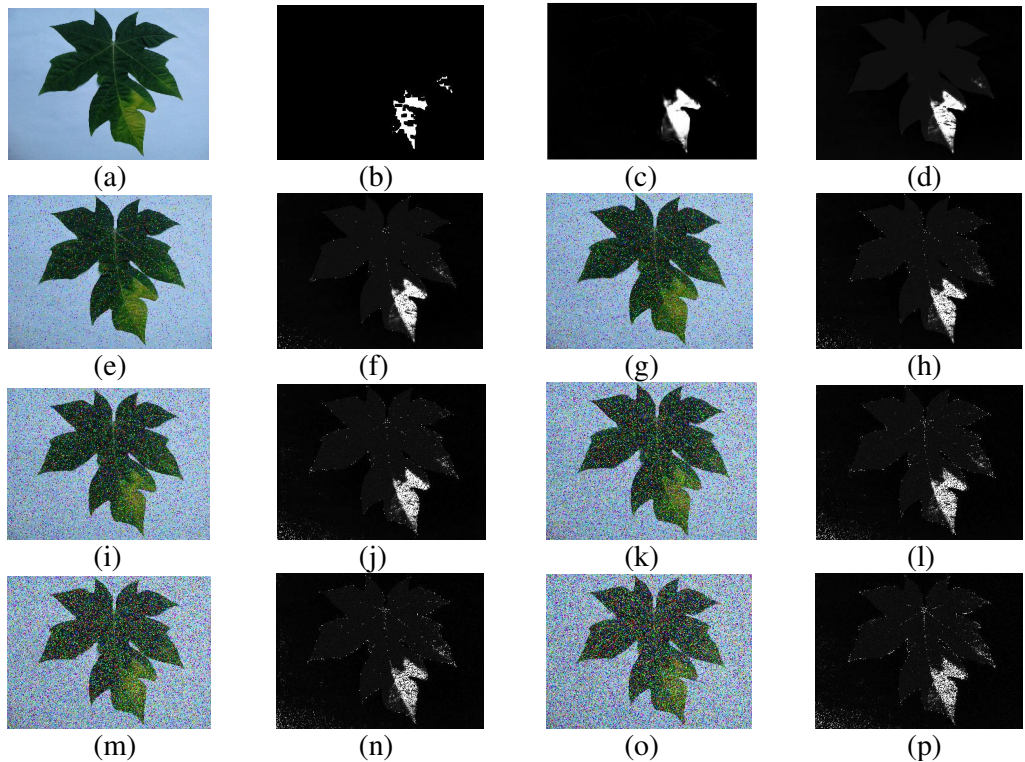

Figure 5. Results of the image segmentation with ANN: (a) source image without noise; (b) goal image; (c) segmentation of (a) with ANN and training without noise (Table 1 number 13); (d) segmentation of (a) and training with noise ; (e) source image with noise of 5%; (f) segmentation of (e) with ANN; (g) noise of 10%; (h) segmentation of (g) with ANN; (i) noise of 15%; (j) segmentation of (i) with ANN ; (k) noise of 20%; (l) segmentation of (k) with ANN; (m) noise of 25%; (n) segmentation of (m) with ANN; (o) noise of 30%; (p) segmentation of (o) with ANN.

## ACKNOWLEDGEMENTS

The authors would like to thank the support of UNEMAT, UNESP, LPSSD, CAPES, FAPEMAT - process (344915/2012) and CNPQ - process (309023/2012-2).